\documentclass[conference]{IEEEtran}
\IEEEoverridecommandlockouts
\usepackage{cite}

\usepackage{comment}
\usepackage[normalem]{ulem}
\usepackage{amsmath,amssymb,amsfonts}
\usepackage{algorithmic}
\usepackage{algorithm}
\usepackage{graphicx}
\usepackage{textcomp}
\usepackage{xcolor}

\newtheorem{lem}{Lemma}[section]
\newtheorem{theo}{Theorem}[section]

\newtheorem{rem}{Remark} 

\usepackage{dsfont}

\usepackage{epstopdf}

\newtheorem{ass}{Assumption}

\newcommand{\carre}     {\hfill$\Box$}

\def\BibTeX{{\rm B\kern-.05em{\sc i\kern-.025em b}\kern-.08em
    T\kern-.1667em\lower.7ex\hbox{E}\kern-.125emX}}
\begin{document}

\title{Online Optimization for Network Resource Allocation and Comparison with Reinforcement Learning Techniques
\thanks{This work was supported by a grant from the Natural Sciences and Engineering Research Council of Canada and Ericsson Canada. }
}

\author{\IEEEauthorblockN{ Ahmed Sid-Ali\IEEEauthorrefmark{1}, Ioannis Lambadaris\IEEEauthorrefmark{2}, Yiqiang Q. Zhao\IEEEauthorrefmark{1}, Gennady Shaikhet\IEEEauthorrefmark{1}, and Amirhossein Asgharnia\IEEEauthorrefmark{2}}
\IEEEauthorblockA{\IEEEauthorrefmark{1}School of Mathematics and Statistics
\\}
\IEEEauthorblockA{\IEEEauthorrefmark{2}Department of Systems and Computer Engineering\\
Carleton University, Ottawa, Ontario\\
Emails: Ahmed.Sidali@carleton.ca; \{ioannis,amirhosseinasgharnia\}@sce.carleton.ca; \{zhao,gennady\}@math.carleton.ca}

}

\maketitle

\begin{abstract}
We tackle in this paper an online network resource allocation problem with job transfers. The network is composed of many servers connected by communication links. The system operates in discrete time; at each time slot, the administrator reserves resources at servers for future job requests, and a cost is incurred for the reservations made. Then, after receptions, the jobs may be transferred between the servers to best accommodate the demands. This incurs an additional transport cost. Finally, if a job request cannot be satisfied, there is a violation that engenders a cost to pay for the blocked job. We propose a randomized online algorithm based on the exponentially weighted method. We prove that our algorithm enjoys a sub-linear in time regret, which indicates that the algorithm is adapting and learning from its experiences and is becoming more efficient in its decision-making as it accumulates more data. Moreover, we test the performance of our algorithm on artificial data and compare it against a reinforcement learning method where we show that our proposed method outperforms the latter.

\end{abstract}


\section{Introduction}
Online optimization is a framework where a decision maker sequentially chooses decision variables over time to minimize the sum of a sequence of loss functions. The decision maker does not have full access to the data at once but receives it incrementally over time. Moreover,  the data source is viewed as arbitrary, and thus, only the empirical properties of the observed data sequence matter. This allows for example to address the dynamic variability of the traffic requests at modern communication networks. Online optimization has further applications in a wide range of fields; see, e.g. \cite{b12, b13, b14, b15}.
Moreover, given that the decision maker has only access to limited/partial information, globally optimal solutions are in general not realizable. Instead, one searches for algorithms that perform relatively well compared to the overall {\it ideal} best static solution in hindsight which has full access to the data. This performance metric is referred to as \textit{regret} in the literature; see, e.g. \cite{b33} for an overview. In particular, if an algorithm incurs regret that increases sub-linearly with time, it demonstrates its ability to make increasingly better decisions over time despite the inherent challenges of limited information and uncertainty in the online setting.

 In this paper, we tackle the problem of resource reservation in communication networks with job transfer. In particular, we consider a network composed of many servers where an administrator reserves resources at each server to meet future job requests that arrive at each server sequentially. Then, the jobs can be transferred between the servers to best accommodate the demand. The reservation of jobs together with their transfer incur specific costs. Moreover, a violation cost is incurred for each unprocessed job. The problem is then to minimize the cumulative costs. However, given the online nature of the problem, one searches for an online algorithm that behaves relatively well in comparison to the best solution in hindsight. In particular, we propose a randomized algorithm based on the exponentially weighted method. We first prove a sub-linear upper bound for the regret and then test our algorithm on simulated data. In particular, we propose a scheme that allows us to solve the underlying optimal transfer optimization problem in an online fashion. Finally, we compare our procedure against a reinforcement learning algorithm where we show that we obtain better results.

\section{Online resource reservation in general network}  
\label{model-sec} 
\begin{figure}
    \centering
    \includegraphics[scale=1.0]{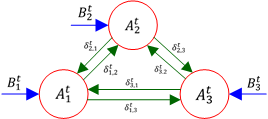}
    \caption{The network model for three nodes at time $t$}
    \label{fig:model}
\end{figure}

Consider a network composed of $N$ servers connected by communication links. The network provides access to computing resources for clients. For simplicity, we assume that there is a single type of resource (e.g. memory, CPU, etc.). Denote by $\mathcal{A}_n$ the total number of resources available at the $n$-th server. The system operates in discrete time slots $t=1,2,\ldots$ where, at each $t$, the following process takes place: 
\begin{itemize} 
\item \textbf{Resource reservation}: the network administrator selects the resources $A^t=(A^t_1,\ldots,A^t_N)$ to make available at each server.  

\item \textbf{Job requests}: the network receives job requests $B^t=(B^t_1,\ldots,B^t_N)$ from its clients to its servers. 
\item \textbf{Job transfer}: the network administrator can shift jobs between the servers to best accommodate the demand.  
\end{itemize} 
Let $\delta^t_{n,m}$ be the number of jobs transferred from the server $n$ to the server $m$ at time slot $t$. Not that these coefficients depend on $A^t$ and $B^t$. However, to keep the notation simple, in the sequel we suppress this dependency. Fig. \ref{fig:model} shows the network for two nodes.

Suppose that the reservations, the job transfers, and the violations incur costs defined as follows:
\paragraph*{\textbf{-Reservation cost}} 
\begin{align*}
C_R(A^t)=\sum_{n=1}^N  f^R_n(A^t_n).
\end{align*} 
\paragraph*{\textbf{-Violation cost}} incurred  when certain job requests cannot be satisfied  
\begin{align*} 
C_V(A^t,B^t)=\sum_{n=1}^N f^V_n\big(B_n^t-A_n^t-\sum_{m=1}^N\delta^t_{n,m}\big).
\end{align*} 
\paragraph*{\textbf{-Transfer cost}} incurred by the transfer of jobs from the peripheral servers to the central server 
\begin{align*}
&C_T(A^t,B^t)=\sum_{n=1}^N\sum_{m\neq n} f^T_{n,m}\big(\delta^t_{n,m}\big). 
\end{align*} 
Here, $f^R_n,f^V_n,f^T_{n,m}$, for $1\leq n\leq N$ are some positive functions. Therefore, after receiving the job requests $B^t$ at time $t$, the job transfers coefficients $\{\delta^t_{n,m},1\leq n,m \leq N\}$ are solutions to the following offline minimization problem: 
 \begin{equation}  
 \begin{split}
 \begin{tabular}{l}
$\min \sum\limits_{n=1}^N\bigg(\sum\limits_{m\neq n}f^T_{n,m}(\delta^t_{n,m})+f^V_n\big(B_n^t-A_n^t-\sum\limits_{m\neq n}\delta^t_{n,m}\big)\bigg),$\\
s.t.: $\delta^t_{n,m}\leq\min\big\{ (B^t_n-A_n^t)^+,(A^t_{m}-B^t_{m})^+ \big\},\forall  n\neq m$ 
 \end{tabular}
\label{delta-opt-prob}
 \end{split}
 \end{equation} 

The goal is then to minimize, at each time slot $t$ the total cost $C(A^t,B^t):=C_R(A^t)+C_V(A^t, B^t)+C_T(A^t, B^t)$ of reservation, violation, and transport cost, and thus to solve
 \begin{equation} 
\begin{split}
\begin{tabular}{l}
$\min\limits_{A^t\in\mathcal{A}} C(A^t,B^t)$. 
\end{tabular}
\label{opt-pb-off}
\end{split}
\end{equation}
Nevertheless, since the reservations $A^t$ are selected {\it before} the job requests $B^t$ are received, then the optimal solution for $(\ref{opt-pb-off})$ is out of reach. Therefore, one instead aims to solve the following online combinatorial optimization problem:   
 \begin{equation} 
\begin{split}
\min\limits_{\{A^t\}}\sum\limits_{t=1}^T C(A^t,B^t),
\label{opt-pb-on}
\end{split}
\end{equation}
over finite time horizons $T\geq 1$. In particular, one searches for an online control policy that uses the cumulative information available so far to make the reservations at the next time slot. Again, the goal is not to reach the optimal solution to $(\ref{opt-pb-on})$ but instead to obtain a cumulative cost $\sum\limits_{t=1}^T C(A^t,B^t)$ that is not too large compared to some benchmark that knows the job requests in advance. Notice that no assumptions on the statistical properties of the sequence $\{B^t\}_{t\geq 1}$ are made, which makes the classical statistical inference methods inappropriate.

\section{Randomized online algorithm} 
\label{rand-sec}
The classical approach for online optimization is based on the gradient-type algorithms first introduced in \cite{b33}. However, the action set $\mathcal{A}=\prod_{n=1}^N\mathcal{A}_n$ being a finite set, these methods cannot be applied in the current setting. Moreover, no sublinear regret is possible in a general adversarial setting using deterministic algorithms; see, e.g. \cite[Chapter 4]{b22} for more details. The idea is thus to add randomization in the decision process such that, at every time slot $t$, the network administrator draws the reservation $A^t$ randomly from a probability distribution $P^t$, which is in turn updated once the job requests are received. Then, the reservation $A^t$ becomes a random variable with probability distribution $P^t$. Denote by $\mathcal{P}(\mathcal{A})$ the space of probability distributions over the reservations set $\mathcal{A}$ identified by the standard simplex 
\begin{align*}   
\mathcal{P}(\mathcal{A})=\bigg\{P=(p_a)_{a\in\mathcal{A}}\in\mathbb{R}_+^{\prod_{n=1}^N\mathcal{A}_n}:\sum\limits_{a\in\mathcal{A}}p_{a}=1\bigg\}. 
\end{align*} 
 The expected reservation cost is then defined by 
\begin{align*}
\mathbb{E}_{P^t}[C_R(A^t)]=\sum_{a\in\mathcal{A}}p^t_{a} C_R(a).
\end{align*}
Similarly, the conditional expected transfer and violation costs given that the job request $B^t=b^t$ are defined by
\begin{align*}
\mathbb{E}_{P^t}[C_T(A^t,b^t)]=\sum_{a\in\mathcal{A}}p^t_{a}C_T(a,b^t),
\end{align*}
and 
\begin{align*}
\mathbb{E}_{P^t}[C_V(A^t,b^t)]=\sum_{a\in\mathcal{A}}p^t_{a}C_V(a,b^t).
\end{align*}  
Therefore, the conditional expectation of the total cost, with respect to a probability distribution $P\in\mathcal{P}(\mathcal{A})$, given that the job requests vector $B^t=b^t$ is $\mathbb{E}_{P}[C(A^t,b^t)].$ The goal is then to find a randomized control policy that produces a sublinear regret with a high probability.
 
\section{Exponentially weighted algorithm}   
\label{algo-sec}
We propose an online randomized algorithm based on the exponentially weighted method (see e.g. \cite[Chapter 2]{b33}). The idea is to assign a weight to each reservation vector based on its past performances, and then take the corresponding convex combination of the vertices as the probability distribution $P^t$ at time slot $t$. In particular, for any $a\in\mathcal{A}$, 
\begin{equation}
\begin{split}
P^t_a=\frac{w^t(a)}{\sum_{a\in\mathcal{A}}w^t(a)}, 
\end{split}
\end{equation}
where the weights are given as 
\begin{equation}
\begin{split}
w^t(a)=\exp\bigg\{-\eta \sum_{s=1}^{t-1}C(a,b^s) \bigg\},
\end{split}
\end{equation}
with $\eta>0$ a positive parameter, and $b^1,\ldots,b^{t-1}$ are the values of job requests observed so far. Here $w^1(a)=1$ for all $a\in\mathcal{A}$. Thus we initialize the algorithm with a uniform distribution. Also, the following recursion is easy to see 
 \begin{align}
 w^t(a)=w^{t-1}(a)\exp\bigg\{-\eta C(a,b^{t-1}) \bigg\}. 
 \label{rec-exp}
 \end{align}

\subsection{Regret bound}
 Let $\{P^t\}_{t\geq 1}$ be the sequence of probability distribution given by the exponentially weighted strategy and let $\{A^t\}_{t\geq 1}$ be the sequence of random reservations generated according to $\{P^t\}_{t\geq 1}$. We define the \textit{regret} over a finite horizon $T>0$ as 
\begin{equation}
\label{eq:regret}
R_T=\sum_{t=1}^T C(A^t,B^t)-\min_{a\in\mathcal{A}}\sum_{t=1}^TC(a,B^t). 
\end{equation}
The regret quantifies the difference between the cumulative cost incurred by the exponentially weighted algorithm and the cumulative cost that would have been experienced if the algorithm had possessed perfect information and could have made the best-fixed decisions in hindsight. In particular, we show next that, under boundedness assumptions, the regret, which is here a random variable, is sublinear in $T$ with high probability.    
\begin{ass} There exists a constant $\Theta>0$ such that, for all $a,b\in\mathcal{A}$,
\begin{align*}
\big|C_R(a)\big|\leq \Theta,\big|C_T(a,b)\big|\leq \Theta,\mbox{ and } \big|C_V(a,b)\big|\leq \Theta. 
\end{align*}
\label{ass}
\end{ass}
  
\begin{theo}
Take $\eta=\sqrt{\frac{\log|\mathcal{A}|}{T}}$. Therefore, under the Assumption \ref{ass}, for any $0<\delta<1$, the regret associated with the Exponentially weighted strategy over finite horizon $T>0$ satisfies, with probability at least $1-\delta$, 
\begin{align}
R_T\leq  \bigg(\frac{9\Theta^2}{8}+1\bigg)\sqrt{T\log |\mathcal{A}|}+3\Theta\sqrt{\frac{1}{2} \log (\delta^{-1}) T }. 
\label{reg-bound} 
\end{align}
\label{reg-theo}
\end{theo}
The proof of Theorem \ref{reg-theo} is based on the following lemma that establishes an upper-bound for the difference between the cumulative expected cost w.r.t. the sequences $\{P^t\}_{t\geq 1}$ and the expected cumulative cost obtained by an optimal fixed probability distribution in hindsight. 
\begin{lem}
Under Assumption \ref{ass}, the sequence $\{P^t\}_{t\geq 1}$ of probability distributions engendered by the exponentially weighted update satisfies, for all $T\geq 1$, 
\begin{equation}
\begin{split}
\sum_{s=1}^T \mathbb{E}_{P^s}[C(A^s,b^s)]- \inf_{P\in\mathcal{P}(\mathcal{A})}&\sum_{s=1}^T \mathbb{E}_P[C(A,b^s)]\\
&\leq  \frac{9\eta\Theta^2}{8}T+\frac{1}{\eta} \log |\mathcal{A}|,
\label{reg-exp}
\end{split}
\end{equation}
for any sequence $\{B_t\}_{t\geq 1}=\{b_t\}_{t\geq 1}$ of job requests.  In particular, taking $\eta=\sqrt{\frac{\log|\mathcal{A}|}{T}}
$ gives
\begin{equation}
\begin{split}
\sum_{s=1}^T \mathbb{E}_{P^s}[C(A^s,b^s)]- \inf_{P\in\mathcal{P}(\mathcal{A})}&\sum_{s=1}^T \mathbb{E}_P[C(A,b^s)]\\
&\leq  \bigg(\frac{9\Theta^2}{8}+1\bigg)\sqrt{T\log |\mathcal{A}|}. 
\end{split}
\label{reg-exp2}
\end{equation}
\label{reg-exp-theo}
\end{lem}

\paragraph*{Proof of Lemma \ref{reg-exp-theo}} First, notice that 
\begin{align*}
\log\frac{\sum_{a\in\mathcal{A}}w^{t+1}(a)}{\sum_{a\in\mathcal{A}}w^1(a)}&=\log \sum_{a\in\mathcal{A}}w^{t+1}(a)-\log |\mathcal{A}|\\
&\geq -\eta \min_{a\in\mathcal{A}}\sum_{s=1}^t C(a,b^s)-\log |\mathcal{A}|.
\end{align*}
On the other hand, by the log function properties in the telescopic sum one gets  
\begin{align*}
\log \frac{\sum_{a\in\mathcal{A}}w^{t+1}(a)}{\sum_{a\in\mathcal{A}}w^1(a)}=\sum_{s=1}^t\log\frac{\sum_{a\in\mathcal{A}}w^{s+1}(a)}{\sum_{a\in\mathcal{A}}w^{s}(a)}.
\end{align*}
Moreover, define the random variable $A^s$ such that $\mathbb{P}(A^s=a)=\frac{w^s(a)}{\sum_{a\in\mathcal{A}}w^{s}(a) }$. Thus, by $(\ref{rec-exp})$, one obtains
\begin{align*}
&\log\frac{\sum_{a\in\mathcal{A}}w^{s+1}(a)}{\sum_{a\in\mathcal{A}}w^{s}(a)}=\log\mathbb{E}_{P^s}\big[\exp\big\{-\eta C(A^s,b^s) \big\} \big]. 
\end{align*}
By Hoeffding's lemma (see, e.g. \cite[Lemma 2.6]{b31}), one gets 
\begin{align*}
\log\frac{\sum_{a\in\mathcal{A}}w^{s+1}(a)}{\sum_{a\in\mathcal{A}}w^{s}(a)}&\leq -\eta\mathbb{E}_{P^s}[C(A^s,b^s)]+\frac{9\eta^2\Theta^2}{8}.
\end{align*} 
Therefore, 
\begin{align*}
-\eta \inf_{a\in\mathcal{A}}&\sum_{s=1}^T C(a,b^s)-\log |\mathcal{A}|\\
&\leq \sum_{s=1}^T (-\eta\mathbb{E}_{P^s}[C(A^s,b^s)]+\frac{9\eta^2\Theta^2}{8}). 
\end{align*}
Finally, notice that by linearity 
\begin{align*}
\inf_{a\in\mathcal{A}}\sum_{s=1}^T C(a,b^s)=\inf_{P\in\mathcal{P}(\mathcal{A})}\sum_{s=1}^T \mathbb{E}_P[(A,b^s)],
\end{align*}
from which we get $(\ref{reg-exp})$. Taking $\eta=\sqrt{\frac{\log|\mathcal{A}|}{T}}$ gives $(\ref{reg-exp2})$.\carre

\paragraph*{Proof of Thereom \ref{reg-theo}} Suppose that the values of job requests are $\{B^t\}_{t=1}^T=\{b^t\}_{t=1}^T$. Then, let $A^*$ be the optimal solution to the following combinatorial optimization problem  
 \begin{equation} 
\begin{split}
\begin{tabular}{l}
$\min\limits_{A\in\mathcal{A}} \sum\limits_{t=1}^T C(A,b^t)$.
\end{tabular}
\end{split}
\label{opt-pb-hind-no-rand}
\end{equation}
 Therefore, the regret of not playing $A^*$ over the horizon $T$ is given by
\begin{align}
R_T= \sum_{t=1}^T \big( C(A^t,b^t)- C(A^*,b^t)\big). 
\label{reg-non-rand}
\end{align}
Define the random variables 
\begin{align*}
X^t= C(A^t,b^t)-\mathbb{E}_{P^t}[C(A^t,b^t)]. 
\end{align*}
Then, by the Assumption \ref{ass}, $|X^t|\leq 3 \Theta$. Therefore, $\{X_t\}_{t\geq 1}$ is a sequence of bounded martingales differences, and $Y_T=\sum_{t=1}^TX_t$ is a martingale with respect to the filtration $\mathcal{F}_T=\sigma(A^t,t\leq T)$. Thus, by Hoeffding-Azuma's inequality \cite{b32} one gets that, for any $0<\delta<1$,    
\begin{align*}
\mathbb{P}\bigg(Y_T\leq \sqrt{\frac{1}{2} \log (\delta^{-1}) T 9\Theta^2}\bigg)\geq 1-\delta.
\end{align*}  
Thus, with probability at least $1-\delta$, one has 
 \begin{align*}
\sum_{t=1}^T C(A^t,b^t) \leq \sum_{t=1}^T \mathbb{E}_{P^t} [C(A^t,b^t)]+\sqrt{   \frac{1}{2} \log (\delta^{-1}) T 9\Theta^2}. 
\end{align*} 
Now, define the probability distribution $P_*\in\mathcal{P}(\mathcal{A})$ as the optimal solution to the following optimization problem    
\begin{equation}
\begin{split}
\min\limits_{P\in\mathcal{P}(\mathcal{A})}\sum\limits_{t=1}^T \mathbb{E}_{P}[C(A,b^t)],
\label{p-star-stat-K}
\end{split}
\end{equation}   
 if the values of job requests $\{B^t\}_{t=1}^T=\{b^t\}_{t=1}^T$ were known is advance. Then, since $\delta_{A^*}\in\mathcal{P}(\mathcal{A})$,  $\sum_{t=1}^T\mathbb{E}_{P^*}[C(A,b^t)]\leq \sum_{t=1}^T C(A^*,b^t)$. Therefore, 
 \begin{equation}
 \begin{split}
R_T &\leq \sum_{s=1}^t \mathbb{E}_{P^s}[C(A^s,b^s)]- \inf_{P\in\mathcal{P}(\mathcal{A})}\sum_{s=1}^t \mathbb{E}_P[C(A,b^s)]\\
& \qquad\qquad+3\Theta\sqrt{\frac{1}{2} \log (\delta^{-1}) T }. 
 \end{split}
 \end{equation}
Using Theorem  \ref{reg-exp-theo} gives (\ref{reg-bound}). \carre 

\begin{rem}
Theorem \ref{reg-theo} indicates that $(R_T/ T)\rightarrow 0$ in the limit $T\rightarrow\infty$ with an arbitrarily large probability. Thus, as the algorithm processes more data through time slots, the regret does not increase proportionally with the number of time slots but rather at a slower rate. In other words, the exponentially weighted algorithm improves its performance over time. 
\end{rem}

\section{Numerical experiments} 
We propose now to test the performance of the exponentially weighted algorithm on simulated data and compare it with a reinforcement learning algorithm. Notice that, implementing our algorithm requires the evaluation, at each slot $t$, of the cost functions $C(a,b^s)$ for each $a\in\mathcal{A}$ and all $b^1,\ldots, b^{t-1}$ observed so far, and thus solving the problem (\ref{delta-opt-prob}) for each combination $(a,b^s)$. This, however, can be infeasible in practice for large network size $N$ and reservation set $\mathcal{A}$. To remedy, we propose in Section \ref{on-job-tran} an exploration strategy where one evaluates the costs for only some new vectors $a\in\mathcal{A}$ at each time slot randomly chosen and then progressively explores new combinations through time slots. Then, we compare our results with a reinforcement learning method described in Section \ref{reinf-sec}. 

\subsection{Exact optimization for the job transfer problem}
First, suppose that one can solve the optimal job transfer problem $(\ref{delta-opt-prob})$ for all combinations of reservations and job request vectors. In particular, at time slot $t$, after observing the job request $B^t=b^t$, one evaluates the cost $C(a,b^{t-1})$ for all $a\in\mathcal{A}$, and then stores the results in memory and used to construct the probability distribution $P^t$. This procedure is detailed in Algorithm \ref{al:alg1}.

\begin{algorithm}
\caption{Resource Reservation with Exact job transfer}
 At each time slot $t$
\begin{itemize}
\item For each $a\in\mathcal{A}$: 
 \begin{itemize}
 \item Retrieve $C(a,b^s)$ from memory for all $1\leq s\leq t-1$ 
  \item Construct the probability distribution $P^t$ as
 \end{itemize}
 \begin{equation*}
w^t(a)=\exp\bigg\{-\eta \sum_{s=1}^{t-1}C(a,b^s) \bigg\},\quad P^t_a=\frac{w^t(a)}{\sum\limits_{a\in\mathcal{A}}w^t(a)} 
\end{equation*}
\item Select $a^t$ randomly from $P^t$ 
\item Observe $b^t$ and pay the cost $C(a^t,b^t)$
 \item Compute $C(a,b^{t})$ for each $a\in\mathcal{A}$ by solving problem $(\ref{delta-opt-prob})$ exactly and store it in memory 
\end{itemize}
\label{al:alg1}
\end{algorithm} 

As a modification to Algorithm \ref{al:alg1}, we added a discount factor to the weight update equation. The weight update equation will be as follows,

\begin{equation}
    \label{eq:w_update_new}
    w^t(a)=\exp\bigg\{-\eta \sum_{s=1}^{t-1} W^{t-s} C(a,b^s) \bigg\}.
\end{equation}

\noindent In \eqref{eq:w_update_new}, $W$ is a discount factor. The discount factor puts more emphasis on the recent experiences. Thus, the modified algorithm has a more robust behaviour in non-deterministic environment.

\subsection{Random exploration}
\label{on-job-tran}

Suppose now that at each time interval $[t,t+1]$, a maximum of $K(N)$ cost functions can be evaluated. Notice that there are a maximum of $(t-1)\cdot |\mathcal{A}|$ cost functions to evaluate at each time slot $t$ to build the probability distribution $P^t$. Therefore, given this computational constraint, we propose to evaluate the cost $C(a,b^s)$ for only $K(N)$ combinations of $a\in \mathcal{A}$ and $s=1,\ldots,t$ randomly selected and then store these combinations in a database, say $S^t$ together with the corresponding costs. Then, in the next time slot, we evaluate these costs for new $K(N)$ combinations and so on. This progressive exploration of new combinations makes the weights $w^t(a)$ progressively more accurate. This procedure is detailed in Algorithm \ref{algo2}.

\subsection{Reinforcement Learning}
\label{reinf-sec}
To show the effectiveness of our method, we compare it with a reinforcement learning (RL) approach. The RL algorithms are mostly based on Markov Decision Processes (MDP) \cite{b25}. Formally, a MDP can be represented by a tuple $(S,A,T,G,\pi)$, where $S$ represents the set of states, $A$ is the set of actions, $T$ is the transition probability, $G$ is the reward function which returns the reward of action in a certain state, and $\pi$ is the policy that defines the agent's strategy or behavior specifying the action to take in each state. Moreover, in an MDP, once an action is taken, the agent moves from its state $s$ to a new state $s'$ and the environment returns a reward. Notice however that in the resource reservation problem considered in this paper, there is a single state. An adequate RL problem without state transition is the $n$-armed bandit problem \cite{b25}.

In an $n$-armed bandit problem, an agent repeatedly faces a choice among $n$ different options, or actions. After each action, a reward $G$ is received as a score given by the environment to the agent. The goal is to select actions $a$ to maximize the reward $G$ over time. Of course, the rewards are not known in advance, we thus rather estimate the expected toward for each action. The estimations of the expected  rewards are calculated iteratively as follows:
\begin{equation}
    \label{eq:update}
    Q(a) \leftarrow Q(a) + \beta (G - Q(a)),
\end{equation}
 where $Q(a)$ is the estimation of expected reward by taking action $a$, $G$ the reward signal, $\beta$ the learning rate. Therefore, one selects the actions randomly according to a probability distribution constructed using the \textit{softmax} function (see \cite{b27}) as follows: 
\begin{equation}
    \label{eq:softmax}
    P^t_a = \frac{\text{exp}(\frac{1}{\tau}Q(a))}{\sum\limits_{a\in \mathcal{A}} \text{exp}(\frac{1}{\tau}Q(a))},
\end{equation}
where $P^t_a$ is the probability of action $a$ being selected at time slot $t$. The term $\tau$ is the \textit{softmax temperature} which adjusts action selection. It should be mentioned that in Eq. (\ref{eq:softmax}) all the elements of the value vector are less than or equal to zero.

\begin{algorithm}[H]
\caption{Resource Reservation with random exploration}
 At each time slot $t$
 \begin{itemize}
 \item For each $a\in\mathcal{A}$
 \begin{itemize}
     \item For each $1\leq s\leq t-1$
     \begin{itemize}
         \item if $(a,b^s)\in S^{t-1}$: retrieve $C(a,b^s)$ from memory and set
         \begin{align*}
             \Tilde{c} (a,b^s)= C(a,b^s)
         \end{align*}
         \item else: select randomly $\zeta$ from $[0,3\Theta]$ and set 
         \begin{align*}
             \tilde{c} (a,b^s)= \zeta
         \end{align*}
     \end{itemize}
 \end{itemize}
  \item Construct the probability distribution $P^t$ as
\begin{align*}
w^t(a)=\exp\bigg\{-\eta \sum_{s=1}^{t-1}\tilde{c}(a,b^s) \bigg\},\quad P^t_a=\frac{w^t(a)}{\sum\limits_{a\in\mathcal{A}}w^t(a)}
\end{align*}
\item Select randomly $a^t$ from $P^t$ and observe $b^t$
\item Evaluate and pay the current cost $C(a^t,b^t)$
\item Select randomly $K(N)-1$ combinations $(a,b^s)\notin S^{t-1}\cup (a^t,b^t)$ with $a\in\mathcal{A}$ and $1\leq s\leq t$ and evaluate the corresponding costs $C(a,b^s)$
  \item Update $S^{t-1}\rightarrow S^t$ by adding the new evaluated combinations
\end{itemize}
\label{algo2}
\end{algorithm}

To solve the reservation problem with reinforcement learning each action $a$ represents the reservations for each node. Therefore, we create a $Q$ vector that has as many elements as the reservation set $\mathcal{A}$. At each time step, an action $a$ is then selected with probability $P^t_a$, given by \ref{eq:softmax}. After selecting $a$, the request $b$ is observed and the reward is calculated. The expected reward of action $a$ is then updated using (\ref{eq:update}).

\begin{figure*}[h]
    \centering
    \includegraphics[scale=0.9]{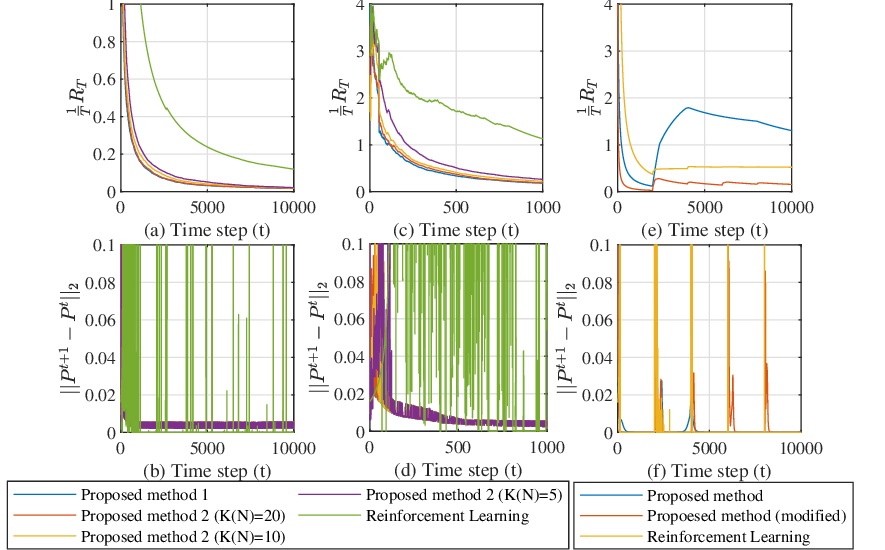}
    \caption{(a) The regret in (\ref{eq:regret}) over the time horizon $T=10,000$. (b) The Euclidean distance between two consecutive $P^t$ over time horizon $T=10,000$. (c) The first 1,000 time steps of (a). (d) The first 1,000 time steps of (b). (e) The regret over 10,000 time steps, with the request changes in every 2,000-time steps. (f) The Euclidean distance between two consecutive $P^t$ over time horizon $T=10,000$.} 
    \label{fig:regret}
\end{figure*}

\subsection{Numerical Simulation}
In this section, we compare our proposed method with the $n$-armed bandit method, as an RL algorithm. We assume that we have three nodes, and each node can have a maximum reservation of five. It is assumed that the minimum reservation must be one. Also, for each node, we will have at least one request. The reservation cost for each node is equal to $f^R=0.5x^2$. The violation cost for each node is equal to $f^V=0.5x^2$. The transfer costs are $f^T_{1,2}=0.2x^2$, $f^T_{1,3}=0.3x^2$, $f^T_{2,1}=0.2x^2$, $f^T_{2,3}=0.2x^2$, $f^T_{3,1}=0.3x^2$, and $f^T_{3,2}=0.2x^2$. Moreover, the step size in the exponentially weighted algorithm is equal to $\eta=\sqrt{\frac{\log(125)}{10000}}$ with a horizon $T=10,000$. The same parameters are used for Algorithm \ref{algo2}. In addition, the computational constraint $K(N)$ is set to $10$, which means that at each time step, the cost function $C(a,b^s)$ is calculated for $10$ possible combinations $(a,b^s)$.

For the RL method, the learning rate $\beta$ is equal to 0.1 in (\ref{eq:update}), and $\tau=0.005$ in (\ref{eq:softmax}). For the RL method, the goal is not minimizing a cost, but it is maximizing a reward. We use the same cost functions as the proposed method, but we multiply them by $-1$. We used the same request sequence for both methods to conduct a fair comparison. 
In addition, to calculate the optimum reservation for the second term of the regret in (\ref{eq:regret}), an integer genetic algorithm is utilized \cite{b29}.

Fig. \ref{fig:regret} (a) and (c) depict the regret in (\ref{eq:regret}) for both Algorithm \ref{al:alg1} and Algorithm \ref{algo2} and RL on a plot. Fig. \ref{fig:regret} (a) shows that all three methods eventually converge. However, the proposed method 1 and 2 converges much faster than the RL approach. In addition, we tested Algorithm \ref{algo2} with three values of the computaional constraint: $K(N)=5$, $K(N)=10,$ and $K(N)=20$. As the parameter $K(N)$ decreases the simulation time will be less with a slightly less performance. Fig. \ref{fig:regret} (c) is the first 1,000 time steps of Fig. \ref{fig:regret} (a).

Among the three methods, the RL method is faster than Algorithm \ref{al:alg1} and Algorithm \ref{algo2}. The RL method has a run-time of 0.783 seconds. Algorithm \ref{al:alg1} has a run-time of 125.3 seconds, and Algorithm \ref{algo2} has almost the same run-time for $K(N)=5, 10, 20$, which is 64.7 seconds. In other words, for the problem in this paper, Algorithm \ref{algo2} is twice as fast as Algorithm \ref{al:alg1}.

Fig. \ref{fig:regret} (b) compares the Euclidean distance between the probability distribution functions in two consecutive time steps. Fig. \ref{fig:regret} (d) is the magnification of the first 1000 time step of the simulation. The figures show that the probability distribution function of Algorithm \ref{al:alg1} and Algorithm \ref{algo2} converge in less than 500 time steps. However, for the RL method, the convergence does not happen before the time step of 1000.   

Fig. \ref{fig:regret} (e) shows a case, where the job requests are fixed for every 2,000 time steps. At each 2,000 time step, the job request changes and will be the same for the next 2,000 time steps. The figure shows that it takes much longer for the proposed method to get adapted to the new job request. The RL method is more robust than the proposed method; however, it does not decrease the regret over time. The modified proposed method makes the regret converge faster than the other methods. On the other hand, the modified proposed method is more robust and it overcomes changes faster than the other methods. Fig. \ref{fig:regret} (f) compares the Euclidean distance between the probability distribution functions in two consecutive time steps. The modified proposed method will be the subject of our future investigation.    

\section{Extension and Conclusion}
We proposed in this paper a new approach to tackle the problem of optimal resource reservation in communication networks with job transfer. Our algorithm, based on an exponentially weighted approach, enjoys a sub-linear regret in time, which indicates that it is adapting and learning from its experiences, becoming more efficient in its decision-making as it accumulates more data through time slots. In addition, we showed that our algorithm outperforms a reinforcement learning approach on simulated data. 

Many interesting questions remain however open. First, one wonders if a sub-linear bound is possible for the dynamic regret metric. Moreover, one wants to test the algorithm on real-world data. Also, one can investigate more sufficient methods to solve the optimal job transfer problem online, especially for general typologies with a large number of servers. These questions will be the object of future research.


\vspace{12pt}
\color{red}

\end{document}